\documentclass{article}

\usepackage[preprint]{neurips_2019}

\usepackage[utf8]{inputenc} 
\usepackage[T1]{fontenc}    
\usepackage{hyperref}       
\usepackage{url}            
\usepackage{booktabs}       
\usepackage{amsfonts}       
\usepackage{nicefrac}       
\usepackage{microtype}      
\usepackage[dvipsnames]{xcolor}
\usepackage{adjustbox}

\usepackage{tikz}
\usetikzlibrary{calc,positioning,patterns,shapes}

\makeatletter
\def\thanks#1{\protected@xdef\@thanks{\@thanks
        \protect\footnotetext{#1}}}
\makeatother

\title{Torchmeta: A Meta-Learning library for PyTorch}

\author{%
  Tristan Deleu\textnormal{\textsuperscript{1}}\quad
  Tobias W\"{u}rfl\textnormal{\textsuperscript{2}}\quad
  Mandana Samiei\textnormal{\textsuperscript{3}}\\
  \textbf{Joseph Paul Cohen\textnormal{\textsuperscript{1}}\quad
  Yoshua Bengio\textnormal{\textsuperscript{1,\,4,\,5}}}%
  \thanks{
  \textsuperscript{1}\,Universit\'{e} de Montr\'{e}al, %
  \textsuperscript{2}\,Friedrich-Alexander-Universit\"{a}t Erlangen-N\"{u}rnberg, %
  \textsuperscript{3}\,Concordia University, %
  \textsuperscript{4}\,CIFAR Senior Fellow, %
  \textsuperscript{5}\,Canada CIFAR AI Chair. %
  Correspondance: \texttt{tristan.deleu@gmail.com}.}\\
  Mila -- Montreal, Canada\\
}

\begin{document}

\maketitle

\begin{abstract}
  The constant introduction of standardized benchmarks in the literature has helped accelerating the recent advances in meta-learning research. They offer a way to get a fair comparison between different algorithms, and the wide range of datasets available allows full control over the complexity of this evaluation. However, for a large majority of code available online, the data pipeline is often specific to one dataset, and testing on another dataset requires significant rework. We introduce Torchmeta, a library built on top of PyTorch that enables seamless and consistent evaluation of meta-learning algorithms on multiple datasets, by providing data-loaders for most of the standard benchmarks in few-shot classification and regression, with a new meta-dataset abstraction. It also features some extensions for PyTorch to simplify the development of models compatible with meta-learning algorithms. The code is available here: \texttt{\href{https://github.com/tristandeleu/pytorch-meta}{https://github.com/tristandeleu/pytorch-meta}}.
\end{abstract}

\section{Introduction}
\label{sec:introduction}
\vspace*{-0.3em}
Like for any subfield of machine learning, the existence of standardized benchmarks has played a crucial role in the progress we have observed over the past few years in meta-learning research. They make the evaluation of existing methods easier and fair, which in turn serves as a reference point for the development of new meta-learning algorithms; this creates a virtuous circle, rooted into these well-defined suites of tasks. Unlike existing datasets in supervised learning though, such as MNIST \citep{lecun1998gradient} or ImageNet \citep{russakovsky2015imagenet}, the benchmarks in meta-learning consist in datasets of datasets. This adds a layer of complexity to the data pipeline, to the extent that a majority of meta-learning projects implement their own specific data-loading component adapted to their method. The lack of a standard at the input level creates variance in the mechanisms surrounding each meta-learning algorithm, which makes a fair comparison more challenging.

Although the implementation might be different from one project to another, the process by which these datasets of datasets are created is generally the same across tasks. In this paper we introduce Torchmeta, a meta-learning library built on top of the PyTorch deep learning framework \citep{paszke2017automatic}, providing data-loaders for most of the standard datasets for few-shot classification and regression. Torchmeta uses the same interface for all the available benchmarks, making the transition between different datasets as seamless as possible. Inspired by previous efforts to design a unified interface between tasks, such as OpenAI Gym \citep{openaigym} in reinforcement learning, the goal of Torchmeta is to create a framework around which researchers can build their own meta-learning algorithms, rather than adapting the data pipeline to their methods. This new abstraction promotes code reuse, by decoupling meta-datasets from the algorithm itself.

In addition to these data-loaders, Torchmeta also includes extensions of PyTorch to simplify the creation of models compatible with classic meta-learning algorithms that sometimes require higher-order differentiation \citep{maml17,finn2018learning,rusu2018meta,DBLP:journals/corr/abs-1801-08930}. This paper gives an overall overview of the features currently available in Torchmeta, and is organized as follows: Section~\ref{sec:data-loaders-few-shot-learning} gives a general presentation of the data-loaders available in the library; in Section~\ref{sec:meta-learning-modules}, we focus on an extension of PyTorch's modules called ``meta-modules'' designed specifically for meta-learning, and we conclude by a discussion in Section~\ref{sec:discussion}.

\vspace*{-1em}
\section{Data-loaders for few-shot learning}
\label{sec:data-loaders-few-shot-learning}
\vspace*{-0.5em}
The library provides a collection of datasets corresponding to classic few-shot classification and regression problems from the meta-learning literature. The interface was created to support modularity between datasets, for both classification and regression, to simplify the process of evaluation on a full suite of benchmarks; we will detail this interface in the following sections. Moreover, the data-loaders from Torchmeta are fully compatible with standard data components of PyTorch, such as \texttt{Dataset} and \texttt{DataLoader}. Before going into the details of the library, we first briefly recall the problem setting.

To balance the lack of data inherent in few-shot learning, meta-learning algorithms acquire some prior knowledge from a collection of datasets $\mathcal{D}_{\mathrm{meta}} = \{\mathcal{D}_{1}, \ldots, \mathcal{D}_{n}\}$, called the \emph{meta-training set}. In the context of few-shot learning, each element $\mathcal{D}_{i}$ contains only a few inputs/output pairs $(x, y)$, where $y$ depends on the nature of the problem. For instance, these datasets can contain examples of different tasks performed in the past. Torchmeta offers a solution to automate the creation of each dataset $\mathcal{D}_{i}$, with a minimal amount of problem-specific components.

\vspace*{-0.5em}
\subsection{Few-shot regression}
\label{sec:few-shot-regression}
\vspace*{-0.3em}
A majority of the few-shot regression problems in the literature are simple regression problems between inputs and outputs through different functions, where each function corresponds to a task. These functions are parametrized to allow variability between tasks, while preserving a constant ``theme'' across tasks. For example, these functions can be sine waves of the form $f_{i}(x) = a_{i}\sin(x + b_{i})$, with $a$ and $b$ varying in some range \citep{maml17}. In Torchmeta, the meta-training set inherits from an object called \texttt{MetaDataset}, and each dataset $\mathcal{D}_{i}$ ($i=1,\ldots, n$, with $n$ defined by the user) corresponds to a specific choice of parameters for the function, with all the parameters sampled once at the creation of the meta-training set. Once the parameters of the function are known, we can create the dataset by sampling inputs in a given range, and feeding them to the function.

The library currently contains 3 toy problems: sine waves \citep{maml17}, harmonic function \citep[i.e. sum of two sine waves, ][]{lacoste2018uncertainty}, and sinusoid and lines \citep{finn2018probabilistic}. Below is an example of how to instantiate the meta-training set for the sine waves problem:\\[0.5em]
\begin{adjustbox}{center}
  \begin{tikzpicture}
    \node[font=\footnotesize\ttfamily, text width=\linewidth, inner sep=5pt, fill=CadetBlue!5] at (0, 0) {torchmeta.toy.Sinusoid(num\_samples\_per\_task=\textcolor{OliveGreen}{10}, num\_tasks=\textcolor{OliveGreen}{1\_000\_000}, noise\_std=\textcolor{OliveGreen}{None})};
  \end{tikzpicture}
\end{adjustbox}

\vspace*{-0.5em}
\subsection{Few-shot classification}
\label{sec:few-shot-classification}
\vspace*{-0.3em}
For few-shot classification problems, the creation of the datasets $\mathcal{D}_{i}$ usually follows two steps: first $N$ classes are sampled from a large collection of candidates (corresponding to $N$ in ``$N$-way classification''), and then $k$ examples are chosen per class (corresponding to $k$ in ``$k$-shot learning''). This two-step process is automated as part of an object called \texttt{CombinationMetaDataset}, inherited from \texttt{MetaDataset}, provided that the user specifies the large collection of class candidates, which is problem-specific. Moreover, to encourage reproducibility in meta-learning, every task is associated to a unique identifier (the $N$-tuple of class identifiers). Once the task has been chosen, the object returns a dataset $\mathcal{D}_{i}$ with all the examples from the corresponding set of classes. In Section~\ref{sec:train-test-split}, we will describe how $\mathcal{D}_{i}$ can then be further split into training and test datasets, as is common in meta-learning.

The library currently contains 5 few-shot classification problems: Omniglot \citep{lake2015human,DBLP:journals/corr/abs-1902-03477}, Mini-ImageNet \citep{DBLP:journals/corr/VinyalsBLKW16,ravi2016optimization}, Tiered-ImageNet \citep{ren2018meta}, CIFAR-FS \citep{bertinetto2018meta}, and Fewshot-CIFAR100 \citep{oreshkin2018tadam}. Below is an example of how to instantiate the meta-training set for 5-way Mini-ImageNet:\\[0.5em]
\begin{adjustbox}{center}
  \begin{tikzpicture}
    \node[font=\footnotesize\ttfamily, text width=\linewidth, inner sep=5pt, fill=CadetBlue!5, align=left] at (0, 0) {torchmeta.datasets.MiniImagenet(\textcolor{Brown}{"data"}, num\_classes\_per\_task=\textcolor{OliveGreen}{5}, meta\_train=\textcolor{OliveGreen}{True},\\\hspace*{16em}download=\textcolor{OliveGreen}{True})};
  \end{tikzpicture}
\end{adjustbox}

Torchmeta also includes helpful functions to augment the pool of class candidates with variants, such as rotated images \citep{santoro2016meta}.

\vspace*{-0.2em}
\subsection{Training \& test datasets split}
\label{sec:train-test-split}
\vspace*{-0.2em}
In meta-learning, it is common to separate each dataset $\mathcal{D}_{i}$ in two parts: a training set (or support set) to adapt the model to the task at hand, and a test set (or query set) for evaluation and meta-optimization. It is important to ensure that these two parts do not overlap though: while the task remains the same, no example can be in both the training and test sets. To ensure that, Torchmeta introduces a wrapper over the datasets called a \texttt{Splitter} that is responsible for creating the training and test datasets, as well as optionally shuffling the data. Here is an example of how to instantiate the meta-training set of a 5-way 1-shot classification problem based on Mini-Imagenet:\\[0.5em]
\begin{adjustbox}{center}
  \begin{tikzpicture}
    \node[font=\footnotesize\ttfamily, text width=\linewidth, inner sep=5pt, fill=CadetBlue!5, align=left] at (0, 0) {dataset = torchmeta.datasets.MiniImagenet(\textcolor{Brown}{"data"}, num\_classes\_per\_task=\textcolor{OliveGreen}{5},\\\hspace*{20.8em}meta\_train=\textcolor{OliveGreen}{True}, download=\textcolor{OliveGreen}{True})\\
    dataset = torchmeta.transforms.ClassSplitter(dataset, num\_train\_per\_class=\textcolor{OliveGreen}{1},\\\hspace*{22.3em}num\_test\_per\_class=\textcolor{OliveGreen}{15}, shuffle=\textcolor{OliveGreen}{True})};
  \end{tikzpicture}
\end{adjustbox}
in addition to the meta-training set, most benchmarks also provide a meta-test set for the overall evaluation of the meta-learning algorithm (and possible a meta-validation set as well). These different meta-datasets can be selected when the \texttt{MetaDataset} object is created, with \texttt{meta\_test=True} (or \texttt{meta\_val=True}) instead of \texttt{meta\_train=True}.

\vspace*{-0.2em}
\subsection{Meta Data-loaders}
\label{sec:meta-dataloaders}
\vspace*{-0.2em}
The objects presented in Sections~\ref{sec:few-shot-regression} \& \ref{sec:few-shot-classification} can be iterated over to generate datasets from the meta-training set; these datasets are PyTorch \texttt{Dataset} objects, and as such can be included as part of any standard data pipeline (combined with \texttt{DataLoader}). Nonetheless, most meta-learning algorithms operate better on batches of tasks. Similar to how examples are batched together with \texttt{DataLoader} in PyTorch, Torchmeta exposes a \texttt{MetaDalaoader} that can produce batches of tasks when iterated over. In particular, such a meta data-loader is able to output a large tensor containing all the examples from the different tasks in the batch. For example:\\[0.5em]
\begin{adjustbox}{center}
  \begin{tikzpicture}
    \node[font=\footnotesize\ttfamily, text width=\linewidth, inner sep=5pt, fill=CadetBlue!5, align=left] at (0, 0) {\textcolor{Gray}{\# Helper function, equivalent to Section~\ref{sec:train-test-split}}\\dataset = torchmeta.datasets.helpers.miniimagenet(\textcolor{Brown}{"data"}, shots=\textcolor{OliveGreen}{1}, ways=\textcolor{OliveGreen}{5},\\\hspace*{24.8em}meta\_train=\textcolor{OliveGreen}{True}, download=\textcolor{OliveGreen}{True})\\
    dataloader = torchmeta.utils.data.BatchMetaDataLoader(dataset, batch\_size=\textcolor{OliveGreen}{16})\\[1em]\textcolor{OliveGreen}{for} batch \textcolor{OliveGreen}{in} dataloader:\\\hspace*{1.9em}train\_inputs, train\_labels = batch[\textcolor{Brown}{"train"}] \textcolor{Gray}{\# Size (16, 5, 3, 84, 84) \& (16, 5)}};
  \end{tikzpicture}
\end{adjustbox}

\vspace*{-0.5em}
\section{Meta-learning modules}
\label{sec:meta-learning-modules}
\vspace*{-0.3em}
Models in PyTorch are created from basic components called \emph{modules}. Each basic module, equivalent to a layer in a neural network, contains both the computational graph of that layer, as well as its parameters. The modules treat their parameters as an integral part of their computational graph; in standard supervised learning, this is sufficient to train a model with backpropagation. However some meta-learning algorithms require to backpropagate through an update of the parameters \citep[like a gradient update,][]{maml17} for the meta-optimization (or the ``outer-loop''), hence involving higher-order differentiation. Although high-order differentiation is available in PyTorch as part of its automatic differentiation module \citep{paszke2017automatic}, replacing one parameter of a basic module with a full computational graph (i.e. the update of the parameter), without altering the way gradients flow, is not obvious.

Backpropagation through an update of the parameters is a key ingredient of gradient-based meta-learning methods \citep{maml17,finn2018learning,DBLP:journals/corr/abs-1801-08930,DBLP:journals/corr/abs-1904-03758}, and various hybrid methods \citep{rusu2018meta,DBLP:journals/corr/abs-1810-03642}. It is therefore critical to adapt the existing modules in PyTorch so they can handle arbitrary computational graphs as a substitute for these parameters. The approach taken by Torchmeta is to extend these modules, and leave an option to provide new parameters as an additional input. These new objects are called \texttt{MetaModule}, and their default behaviour (i.e. without any extra parameter specified) is equivalent to their PyTorch counterpart. Otherwise, if extra parameters (such as the result of one step of gradient descent) are specified, then the \texttt{MetaModule} treats them as part of the computational graph, and backpropagation works as expected.

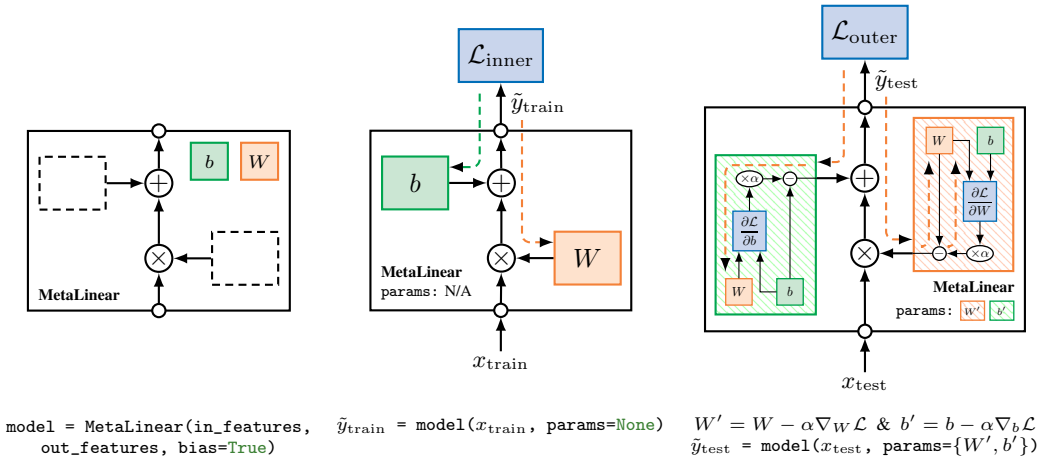
\begin{figure}[t]
  \centering
  \begin{tikzpicture}
    \node[font=\scriptsize, align=center] (code) at (0, 0) {\texttt{model = MetaLinear(in\_features,}\\\texttt{out\_features, bias=\textcolor{OliveGreen}{True})}};
    \node[above=3.1em of code] (figure) {\begin{tikzpicture}
      \begin{scope}[local bounding box=metalinear]
        \node[draw=black, circle, inner sep=0, thick] (times) at (0, 0) {$\times$};
        \node[draw=black, circle, inner sep=0, thick, above of=times] (plus) {$+$};

        \node[inner sep=0, minimum width=1pt, fill=red, fill opacity=0, minimum height=6em, anchor=center] at ($(times)!0.5!(plus)$) {};

        \node[draw=black, rectangle, thick, densely dashed, minimum height=2em, minimum width=2.5em, right=0.5 of times] (W) {};
        \node[draw=black, rectangle, thick, densely dashed, minimum height=2em, minimum width=2.5em, left=0.5 of plus] (b) {};

        \draw[-{latex}, thick] (times) -- (plus);
        \draw[-{latex}, thick] (W) -- (times);
        \draw[-{latex}, thick] (b) -- (plus);

        \node[font=\tiny, inner sep=0, outer sep=0, anchor=south west] at (current bounding box.{south west}) {
          \begin{tikzpicture}{every node/.style={inner sep=0, outer sep=0}}
            \node[anchor=south west] at (0, 0) {\textbf{MetaLinear}};
          \end{tikzpicture}
        };

        \node[font=\scriptsize, inner sep=0, outer sep=0, anchor=north east] at (current bounding box.{north east}) {
          \begin{tikzpicture}
            \node[draw=Orange, fill=Orange!20, rectangle, thick, anchor=north east, minimum size=2em] (W_2) at (0, 0) {$W$};
            \node[draw=Green, fill=Green!20, rectangle, thick, left=5pt of W_2, minimum size=2em] (b_2) {$b$};
          \end{tikzpicture}
        };
      \end{scope}
      \draw[thick] ($(metalinear.{north west})+(-4pt,4pt)$) rectangle ($(metalinear.{south east})+(4pt,-4pt)$);
      \node[inner sep=0, outer sep=0, minimum size=5pt, circle, fill=white, thick, draw=black, anchor=center, yshift=-1pt] (output-interface) at ($(metalinear.north)+(0,5pt)$) {};
      \node[inner sep=0, outer sep=0, minimum size=5pt, circle, fill=white, thick, draw=black, anchor=center, yshift=1pt] (input-interface) at ($(metalinear.south)+(0,-5pt)$) {};

      \draw[-{latex}, thick] (input-interface) -- (times);
      \draw[-{latex}, thick] (plus) -- (output-interface);

    \end{tikzpicture}};
  \end{tikzpicture}\hfill
  \begin{tikzpicture}
    \node[font=\scriptsize, align=center] (code) at (0, 0) {\texttt{$\tilde{y}_{\mathrm{train}}$ = model($x_{\mathrm{train}}$, params=\textcolor{OliveGreen}{None})}\\\texttt{\phantom{p}}};
    \node[above=0.9em of code] (figure) {\begin{tikzpicture}
      \begin{scope}[local bounding box=metalinear]
        \node[draw=black, circle, inner sep=0, thick] (times) at (0, 0) {$\times$};
        \node[draw=black, circle, inner sep=0, thick, above of=times] (plus) {$+$};

        \node[inner sep=0, minimum width=1pt, fill=red, fill opacity=0, minimum height=6em, anchor=center] at ($(times)!0.5!(plus)$) {};

        \node[draw=Orange, fill=Orange!20, rectangle, thick, minimum height=2em, minimum width=2.5em, right=0.5 of times] (W) {$W$};
        \node[draw=Green, fill=Green!20, rectangle, thick, minimum height=2em, minimum width=2.5em, left=0.5 of plus] (b) {$b$};

        \draw[-{latex}, thick] (times) -- (plus);
        \draw[-{latex}, thick] (W) -- (times);
        \draw[-{latex}, thick] (b) -- (plus);

        \node[font=\tiny, inner sep=0, outer sep=0, anchor=south west] at (current bounding box.{south west}) {
          \begin{tikzpicture}{every node/.style={inner sep=0, outer sep=0}}
            \node[anchor=south west] (params) at (0, 0) {\texttt{params:} N/A};
            \node[anchor=south west, yshift=3pt] at (params.{north west}) {\textbf{MetaLinear}};
          \end{tikzpicture}
        };
      \end{scope}
      \draw[thick] ($(metalinear.{north west})+(-4pt,4pt)$) rectangle ($(metalinear.{south east})+(4pt,-4pt)$);
      \node[inner sep=0, outer sep=0, minimum size=5pt, circle, fill=white, thick, draw=black, anchor=center, yshift=-1pt] (output-interface) at ($(metalinear.north)+(0,5pt)$) {};
      \node[inner sep=0, outer sep=0, minimum size=5pt, circle, fill=white, thick, draw=black, anchor=center, yshift=1pt] (input-interface) at ($(metalinear.south)+(0,-5pt)$) {};

      \draw[-{latex}, thick] (input-interface) -- (times);
      \draw[-{latex}, thick] (plus) -- (output-interface);

      \node[above of=output-interface, draw=NavyBlue, fill=NavyBlue!20, thick, minimum height=2em, minimum width=2.5em] (loss) {$\mathcal{L}_{\mathrm{inner}}$};
      \draw[-{latex}, thick] (output-interface) --  node[midway, font=\small, anchor=west] (y) {$\tilde{y}_{\mathrm{train}}$} (loss);
      \node[below=1.3em of input-interface, font=\small, inner sep=2pt, outer sep=0] (x) {$x_{\mathrm{train}}$};
      \draw[-{latex}, thick] (x) -- (input-interface);

      \draw[-{latex}, thick, draw=Orange, rounded corners=3pt, densely dashed] ($(loss.south)+(8pt,-1.4em)$) |- ($(W.west)+(0,6pt)$);
      \draw[-{latex}, thick, draw=Green, rounded corners=3pt, densely dashed] ($(loss.south)+(-8pt,-0.4em)$) |- ($(b.east)+(0,6pt)$);
    \end{tikzpicture}};
  \end{tikzpicture}\hfill
    \begin{tikzpicture}
    \node[font=\scriptsize, align=center] (code) at (0, 0) {$W' = W - \alpha \nabla_{W}\mathcal{L}\ $ \& $\ b' = b - \alpha \nabla_{b}\mathcal{L}$\\\texttt{$\tilde{y}_{\mathrm{test}}$ = model($x_{\mathrm{test}}$, params=$\{W', b'\}$)}};
    \node[above=0em of code] (figure) {\begin{tikzpicture}
      \begin{scope}[local bounding box=metalinear]
        \node[draw=black, circle, inner sep=0, thick] (times) at (0, 0) {$\times$};
        \node[draw=black, circle, inner sep=0, thick, above of=times] (plus) {$+$};

        \node[inner sep=0, minimum width=1pt, fill=red, fill opacity=0, minimum height=4em, anchor=center] at (0, 0) {};

        \node[draw=Orange, rectangle, thick, pattern=north west lines, pattern color=Orange!30, minimum height=2em, minimum width=2.5em, above right=-1.1em and 0.5 of times] (W) {
        \begin{tikzpicture}[scale=0.5, every node/.style={transform shape, line width=0.5pt, solid, inner sep=0, minimum size=0}]
          \node[draw=black, ellipse, fill=white, minimum height=1.2em] (W_times) at (0, 0) {$\times \alpha$};
          \node[above=1.8em of W_times, draw=NavyBlue, fill=NavyBlue!20, minimum width=2.5em, minimum height=3em] (W_dLdW) {$\displaystyle\frac{\partial \mathcal{L}}{\partial W}$};
          \node[left=1.5em of W_times, draw=black, fill=white, circle] (W_sub) {$-$};
          \node[above=7em of W_sub, draw=Orange, fill=Orange!20, minimum size=2em] (W_W) {$W$};
          \coordinate (W_b_dLdW) at ($(W_dLdW.north)+(0.8em,0)$);
          \node[draw=Green, fill=Green!20, minimum size=2em, anchor=center] (W_b) at (W_b_dLdW |- W_W.center) {$b$};

          \draw[-{latex}, solid] (W_dLdW) -- (W_times);
          \draw[-{latex}, solid] (W_times) -- (W_sub);
          \draw[-{latex}, solid] (W_W) -- (W_sub);
          \draw[-{latex}, solid] (W_W) -| ($(W_dLdW.north)+(-0.8em,0)$);
          \draw[-{latex}, solid] (W_b) -- ($(W_dLdW.north)+(0.8em,0)$);

          \draw[-{latex}, densely dashed, draw=Orange, thick, rounded corners=3pt] ($(W_sub.north)+(-1.2em,0)$) -| node[pos=1] (W_head) {} ($(W_sub.north)+(-0.8em,6.5em)$);
          \draw[-{latex}, densely dashed, draw=Orange, thick, rounded corners=3pt] ($(W_sub.north)+(0.8em,0)$) -| ($(W_head)+(2em,0)$);
        \end{tikzpicture}
        };
        \node[draw=Green, rectangle, thick, pattern=north west lines, pattern color=Green!30, minimum height=2em, minimum width=2.5em, below left=-1.3em and 0.5 of plus] (b) {
        \begin{tikzpicture}[scale=0.5, every node/.style={transform shape, line width=0.5pt, solid, inner sep=0, minimum size=0}]
          \node[draw=black, ellipse, fill=white, minimum height=1.2em] (b_times) at (0, 0) {$\times \alpha$};
          \node[below=1.8em of b_times, draw=NavyBlue, fill=NavyBlue!20, minimum width=2.5em, minimum height=3em] (b_dLdb) {$\displaystyle\frac{\partial \mathcal{L}}{\partial b}$};
          \node[right=1.5em of b_times, draw=black, fill=white, circle] (b_sub) {$-$};
          \node[below=7em of b_sub, draw=Green, fill=Green!20, minimum size=2em] (b_b) {$b$};
          \coordinate (b_W_dLdb) at ($(b_dLdb.north)+(-0.8em,0)$);
          \node[draw=Orange, fill=Orange!20, minimum size=2em, anchor=center] (b_W) at (b_W_dLdb |- b_b.center) {$W$};

          \draw[-{latex}, solid] (b_dLdb) -- (b_times);
          \draw[-{latex}, solid] (b_times) -- (b_sub);
          \draw[-{latex}, solid] (b_b) -- (b_sub);
          \draw[-{latex}, solid] (b_b) -| ($(b_dLdb.south)+(0.8em,0)$);
          \draw[-{latex}, solid] (b_W) -- ($(b_dLdb.south)+(-0.8em,0)$);

          \draw[-{latex}, densely dashed, draw=Orange, thick, rounded corners=3pt] ($(b_sub.north)+(1.2em,0.5em)$) -| ($(b_W.{north west})+(0,0.5em)$);
        \end{tikzpicture}
        };

        \draw[-{latex}, thick] (times) -- (plus);
        \draw[-{latex}, thick] (W.west |- times) -- (times);
        \draw[-{latex}, thick] (b.east |- plus) -- (plus);
        \draw[-, thin] (W.west |- times) -- ++(0.75em, 0);
        \draw[-, thin] (b.east |- plus) -- ++(-0.75em, 0);

        \node[font=\tiny, inner sep=0, outer sep=0, anchor=south east] at (current bounding box.{south east}) {
          \begin{tikzpicture}{every node/.style={inner sep=0, outer sep=0}}
            \node[anchor=south east] (params) at (0, 0) {\texttt{params:} \tikz[baseline=0.7ex, scale=0.6]{\node[draw=Orange, pattern=north west lines, pattern color=Orange!30, rectangle, transform shape, minimum height=2em, minimum width=2.5em, line width=0.5pt] {$W'$};} \tikz[baseline=0.7ex, scale=0.6]{\node[draw=Green, pattern=north west lines, pattern color=Green!30, rectangle, transform shape, minimum height=2em, minimum width=2.5em, line width=0.5pt] {$b'$};}};
            \node[anchor=south east, yshift=3pt] at (params.{north east}) {\textbf{MetaLinear}};
          \end{tikzpicture}
        };
      \end{scope}
      \draw[thick] ($(metalinear.{north west})+(-4pt,4pt)$) rectangle ($(metalinear.{south east})+(4pt,-4pt)$);
      \node[inner sep=0, outer sep=0, minimum size=5pt, circle, fill=white, thick, draw=black, anchor=center, yshift=-1pt] (output-interface) at ($(metalinear.north)+(0,5pt)$) {};
      \node[inner sep=0, outer sep=0, minimum size=5pt, circle, fill=white, thick, draw=black, anchor=center, yshift=1pt] (input-interface) at ($(metalinear.south)+(0,-5pt)$) {};

      \draw[-{latex}, thick] (input-interface) -- (times);
      \draw[-{latex}, thick] (plus) -- (output-interface);

      \node[above of=output-interface, draw=NavyBlue, fill=NavyBlue!20, thick, minimum height=2em, minimum width=2.5em] (loss) {$\mathcal{L}_{\mathrm{outer}}$};
      \draw[-{latex}, thick] (output-interface) -- node[midway, font=\small, anchor=west] (y) {$\tilde{y}_{\mathrm{test}}$} (loss);
      \node[below=1.3em of input-interface, font=\small, inner sep=2pt, outer sep=0] (x) {$x_{\mathrm{test}}$};
      \draw[-{latex}, thick] (x) -- (input-interface);

      \draw[-{latex}, thick, draw=Orange, rounded corners=3pt, densely dashed] ($(loss.south)+(8pt,-1.4em)$) |- ($(W.west)+(0,-1.7em)$);
      \draw[-{latex}, thick, draw=Orange, rounded corners=3pt, densely dashed] ($(loss.south)+(-8pt,-0.4em)$) |- ($(b.{north east})+(0,-3pt)$);
    \end{tikzpicture}};
  \end{tikzpicture}
  \caption{Illustration of the functionality of the \texttt{MetaLinear} meta-module, the extension of the \texttt{Linear} module. Left: Instantiation of a \texttt{MetaLinear} meta-module. Middle: Default behaviour, equivalent to \texttt{Linear}. Right: Behaviour with extra parameters \citep[a one-step gradient update,][]{maml17}. Gradients are represented as dashed arrows, in orange for $\partial/\partial W$ and green for $\partial/\partial b$.}
  \label{fig:metalinear}
\end{figure}

Figure~\ref{fig:metalinear} shows how the extension of the \texttt{Linear} module called \texttt{MetaLinear} works, with and without additional parameters, and the impact on the gradients. The figure on the left shows the instantiation of the meta-module as a container for the parameters $W$ \& $b$, and the computational graph with placeholders for the weight and bias parameters. The figure in the middle shows the default behaviour of the \texttt{MetaLinear} meta-module, where the placeholders are substituted with $W$ \& $b$: this is equivalent to PyTorch's \texttt{Linear} module. Finally, the figure on the right shows how these placeholders can be filled with a complete computational graph, like one step of gradient descent \citep{maml17}. In this latter case, the gradient of $\mathcal{L}_{\mathrm{outer}}$ with respect to $W$, necessary in the outer-loop update, can correctly flow all the way to the parameter $W$.

\vspace*{-0.3em}
\section{Discussion}
\label{sec:discussion}
\vspace*{-0.3em}
Reproducibility of data pipelines is challenging. It is even more challenging that some early works, even though they were evaluated on benchmarks that now became classic, did not disclose the set of classes available for meta-training and meta-test (and possibly meta-validation) in few-shot classification. For example, while the Mini-ImageNet dataset was introduced in \citep{DBLP:journals/corr/VinyalsBLKW16}, the split used in \citep{ravi2016optimization} is now widely accepted in the community as the official dataset. The advantage of a library like Torchmeta is to standardize these benchmarks to avoid any confusion.

The other objective of Torchmeta is to make meta-learning accessible to a larger community. We hope that similar to how OpenAI Gym \citep{openaigym} helped the progress in reinforcement learning, with an access to multiple environments under a unified interface, Torchmeta can have an equal impact on meta-learning research. The full compatibility of the library with both PyTorch and Torchvision, PyTorch's computer vision library, simplifies its integration to existing projects. 

Even though Torchmeta already features a number of datasets for both few-shot regression and classification, and covers most of the standard benchmarks in the meta-learning literature, one notable missing dataset in the current version of the library is Meta-Dataset \citep{triantafillou2019meta}. Meta-Dataset is a unique and more complex few-shot classification problem, with a varying number of classes per task. And while it could fit the proposed abstraction, Meta-Dataset requires a long initial processing phase, which would make the automatic download and processing feature of the library impractical. Therefore its integration is left as future work. In the meantime, we believe that Torchmeta can provide a structure for the creation of better benchmarks in the future, and is a crucial step forward for reproducible research in meta-learning.

\section{Acknowledgements}
\label{sec:acknowledgements}
We would like to thank the students at Mila who tested the library, and provided valuable feedback during development. Tristan Deleu is supported by the Antidote scholarship from Druide Informatique.

\bibliographystyle{apalike}
\bibliography{references}

\end{document}